\documentclass[conference]{IEEEtran}

\usepackage[pdftex]{graphicx}
\usepackage{amsmath}
\usepackage{amssymb}
\usepackage{mathtools}

\usepackage{cite}
\usepackage{hyperref}
\usepackage[capitalise, noabbrev]{cleveref}
\usepackage{acronym}

\usepackage{algpseudocode}

\usepackage{array}
\usepackage{flushend}
\usepackage{multirow}
\usepackage{nicematrix}
\usepackage{arydshln}

\usepackage{url}

\newcommand{\crossmat}[1]{\left[#1\right]_\times}

\newcommand{\HatBold}[1]{\hat{\boldsymbol{#1}}}

\newcommand{\eye}[1]{\mathbf{I}_{#1}}
\newcommand{\zeros}[2]{\mathbf{0}_{#1 \times #2}}
\newcommand{\quatRot}[1]{\mathcal{C}(#1)}

\acrodef{MSCKF}{Multi-State Constraint Kalman Filter}
\acrodef{VIO}{visual-inertial odometry}
\acrodef{EKF}{Extended Kalman Filter}
\acrodef{IMU}{inertial measurement unit}
\acrodef{VIMU}{virtual inertial measurement unit}
\acrodef{GPS}{global positioning system}
\acrodef{INS}{inertial navigation system}
\acrodef{RPE}{relative position error}

\IEEEoverridecommandlockouts
\usepackage{tikz}
\usepackage{textcomp}
\usepackage{lipsum}

\newcommand\copyrighttext{%
  \footnotesize \textcopyright 2023 IEEE. Personal use of this material is permitted.
  Permission from IEEE must be obtained for all other uses, in any current or future
  media, including reprinting/republishing this material for advertising or promotional
  purposes, creating new collective works, for resale or redistribution to servers or
  lists, or reuse of any copyrighted component of this work in other works.
  DOI: \href{https://doi.org/10.1109/SDF-MFI59545.2023.10361310}{10.1109/SDF-MFI59545.2023.10361310}}
\newcommand\copyrightnotice{%
\begin{tikzpicture}[remember picture,overlay]
\node[anchor=south,yshift=10pt] at (current page.south) {\fbox{\parbox{\dimexpr\textwidth-\fboxsep-\fboxrule\relax}{\copyrighttext}}};
\end{tikzpicture}%
}

\begin{document}

\title{Online Multi-IMU Calibration Using Visual-Inertial Odometry }

\author{Jacob~Hartzer and Srikanth~Saripalli}

\maketitle
\copyrightnotice

\begin{abstract}
    This work presents a centralized multi-IMU filter framework with online
    intrinsic and extrinsic calibration for unsynchronized inertial measurement units that is robust against
    changes in calibration parameters. The novel \acl{EKF}-based method
    estimates the positional and rotational offsets of the system of sensors
    as well as their intrinsic biases without the use of rigid body geometric constraints.
    Additionally, the filter is flexible in the total number of
    sensors used while leveraging the commonly used \acl{MSCKF} framework for camera measurements.
    The filter framework has been validated using Monte Carlo simulation as
    well as experimentally. In both simulations and experiments, using multiple
    IMU measurement streams within the proposed filter framework
    outperforms the use of a single IMU in a filter prediction step while also
    producing consistent and accurate estimates of initial calibration errors.
    Compared to current state-of-the-art optimizers, the filter produces similar
    intrinsic and extrinsic calibration parameters for each sensor.
    Finally, an open source repository has been provided at
    \url{https://github.com/unmannedlab/ekf-cal} containing both the online
    estimator and the simulation used for testing and evaluation.
\end{abstract}

\IEEEpeerreviewmaketitle

\section{Introduction}

Modern robotic systems utilize a vast array of sensors in localization
which has greatly improved their robustness and accuracy in challenging
environments. Many of these sensors produce feature rich measurements at moderate
rates, which can lead to excellent localization results. At the base of many
\ac{VIO} systems, there is an \ac{IMU} to provide high-rate dead
reckoning estimations of body rates and accelerations. These inertial
measurements are key to the stability of many control algorithms and
improve localization estimation over short distances and times. While generally
the performance of a \ac{VIO} is greatly impacted by the visual processing front and
back end, improvements to the proprioceptive measurement processing can lead to
reductions in drift, improvements in stability, and smoother estimation output.

To ensure high quality inertial measurement data, many ground and airborne
systems use redundant \acp{IMU} to either produce a consensus on measurements, or
simply select the \ac{IMU} that is producing the best measurements. These methods of
inertial measurement, while robust against faults and large baseline distances
between sensors, do not leverage the multiple measurement streams to reduce errors,
but rather discard any of the less-than-ideal measurements. As this reduces
valuable information about the system state, extensive work has been done to
formulate stable and effective methods of filtering multiple \acp{IMU}
measurements that optimally takes advantage of all measurement streams.

\section{Related Work}

The use of multiple \acp{IMU} in a navigation system has been shown to reduce inertial
measurement errors and improve localization accuracy without significantly
increasing computational load \cite{Faizullin,Patel,Bancroft_2011,Zhang}.
Beyond reducing kinematic prediction errors, the use of
multiple \acp{IMU} can produce a more robust and fault-tolerant estimation of
inertial measurements \cite{Egidio,Eckenhoff}.
Two common methods for combining multiple \acp{IMU} are the use of a \ac{VIMU}
and a federated filter architecture.

The \ac{VIMU} observation fusion method probabilistically combines measurements
from all the \acp{IMU} into the functional equivalent of a single sensor, such as in \cite{Huang_2023}.
The \ac{VIMU} fusion method is relatively lightweight computationally and can
easily be incorporated into existing \ac{VIO} systems built around a single
\ac{IMU}. The measurement streams are generally combined using least
squares estimation or an average weighted by the inverse of noise \cite{Larey}.
While computationally efficient, these systems typically require consistent
and synchronized \ac{IMU}, so therefore cannot benefit from ad hoc sensor
additions.

Another framework commonly used are federated filters, which utilize the multiple
\ac{IMU} measurement streams as prediction steps in separate, decentralized filters
\cite{Carlson_federated,Luo_2021}. These stacked filters are then combined
using a single federating filter, which then feeds state parameters and
covariance information back to the stacked filters. Compared to the \ac{VIMU},
the federated filter architecture can combine filters utilizing different
error models and states, which can be very useful if combining sensors of
various quality or calibration. This framework can also more
readily incorporate geometric constraints on the different \acp{IMU}, and
therefore is more appropriate for a system with large baseline distances between
multiple \acp{IMU}. However, like the \ac{VIMU} design, the federated filter
architecture requires synchronization of the stacked filters for information
sharing. Additionally, both systems presume prior calibration of the sensors intrinsic
and extrinsic parameters, which is necessary for these system's accuracy \cite{Wang_2021}.
These calibration parameters are often estimated using state-of-the-art offline
methods, such as \cite{Rehder}. These methods typically require a defined calibration
environment, such as motion in view of a calibration fiducial target.
As it is not always possible to perform these offline calibrations before a system is used
and \ac{IMU} calibration parameters can change with time, it is
advantageous to utilize a framework that can produce online estimates of
calibration parameters while still combining multiple \ac{IMU} measurement streams.

Solving these issues are centralized filter frameworks that simultaneously
estimate intrinsic and extrinsic sensor parameters for multiple \acp{IMU}. In
\cite{Bancroft_2009}, rigid body constraints are applied as a low-noise
measurement within a single centralized filter in order to combine multiple IMU measurement
streams. The noise within this constraint update can be tuned to allow the filter
to adapt to changes in calibration, due to a flexible body or initial uncertainty
in sensor calibration. Extending this, \cite{MIMCVINS} varied the injected noise
of the rigid body constraint with initially large values to allow for large
uncertainties in initial calibration and lower values as time passed as
certainty improved. Effectively, this produces an online estimate of the
\acp{IMU} calibration parameters and is more robust due through its use alongside
camera measurements in an \ac{MSCKF} framework. However, without further system engineering
development, the method of varying injected noise method is not robust to changes in extrinsic parameters as it
is not truly an online estimate of the sensor parameters. As will be discussed
in \cref{sec:obs}, the use of the \ac{MSCKF} framework makes the entire state
observable when intrinsic calibration of all \acp{IMU} is desired, such that online
estimation for all \ac{IMU} parameters is possible.

Therefore, this work proposes a centralized filter framework that is
flexible in the number of sensors used and produces online estimates of each
\ac{IMU}'s intrinsic and extrinsic calibration parameters that can be utilized alongside
camera measurements within a \ac{MSCKF} framework. This
filter provides higher accuracy estimates of body accelerations and angular rates
through the combination of multiple unsynchronized sensors while remaining tolerant of measurement stream
drop outs.

\section{Multi-IMU Filtering}

\begin{figure}
    \centering
    \includegraphics[width = 0.8\linewidth]{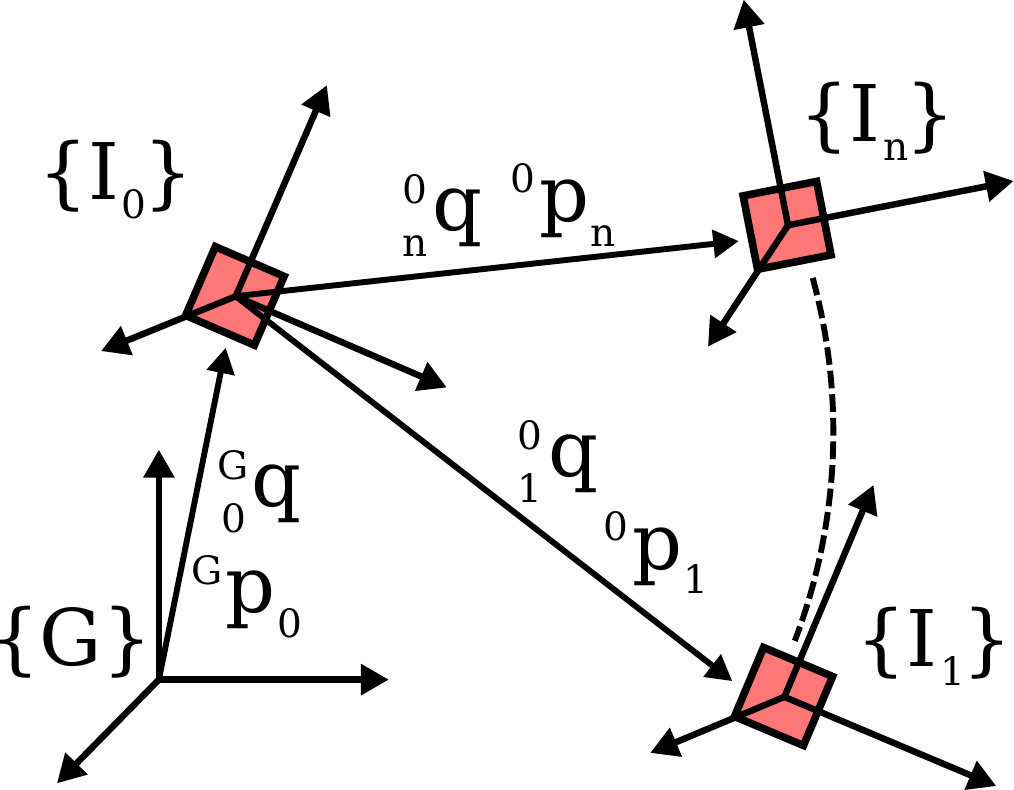}
    \caption{
        Relationship graph between frames of reference showing
        translations and rotations between the primary \{$I_0$\}
        and supplemental \{$I_1$...$I_N$\} \ac{IMU} frames
    }
    \label{fig:setup}
\end{figure}

\subsection{State Formulation}

The filter framework used is based on and extends the \ac{MSCKF},
proposed in \cite{MSCKF}. Typically, the \ac{VIO} filter utilizes a body state
vector containing position, velocity, and angular orientation where the \ac{IMU}
measurements provide the acceleration and angular acceleration estimates for the
sake of kinematic prediction. However, in order to optimally utilize multiple
IMU measurement streams, all measurements are used within the update step of
the filter. As is explained in \cref{sec:udpate} (and explicitly shown in
\cref{eq:pred_measurement}), it is necessary to maintain an estimate
of the current body's acceleration, angular rate, and angular acceleration.
As such, the body state vector is
\begin{equation} \label{eq:base_state}
    \boldsymbol{x}_b =
    \begin{bmatrix}
        \boldsymbol{p}      &
        \boldsymbol{v}      &
        \boldsymbol{a}      &
        \prescript{G}{B}{q} &
        \boldsymbol{\omega} &
        \boldsymbol{\alpha}
    \end{bmatrix}
\end{equation}
where
$\boldsymbol{p}$ is the body position,
$\boldsymbol{v}$ is the body velocity,
$\boldsymbol{a}$ is the body acceleration,
$\prescript{G}{B}{q}$ is the body angular orientation in the global frame,
$\boldsymbol{\omega}$ is the body angular rate, and
$\boldsymbol{\alpha}$ is the body angular acceleration.

If there is uncertainty in the calibration of any of the \acp{IMU} used, it is
possible to extend the filter state to include the calibration parameters and
produce online estimates of the intrinsic and extrinsic values. The calibration
state vector for a single \ac{IMU} $i$ is
\begin{equation} \label{eq:IMU_state}
    \boldsymbol{x}_{I_N} =
    \begin{bmatrix}
        \prescript{B}{}{\boldsymbol{p}}_{I_i} &
        \prescript{B}{I_i}{q}                 &
        \boldsymbol{b}_a                      &
        \boldsymbol{b}_\omega
    \end{bmatrix}
\end{equation}
where
$\prescript{B}{}{\boldsymbol{p}}_{I_i}$ is the \ac{IMU} position in the body frame,
$\prescript{B}{I_i}{q}$ is the \ac{IMU} orientation in the body frame,
$\boldsymbol{b}_a$ is the \ac{IMU} acceleration bias, and
$\boldsymbol{b}_\omega$ is the \ac{IMU} angular rate bias. Note, the body
frame is arbitrary, and leaving it undefined unnecessarily introduces three degrees
of freedom into the system. Therefore in implementation, the body frame $B$
is defined to be coincident with the first \ac{IMU} frame $I_0$. Combining the
body state with the calibration states for $N$ \acp{IMU} gives the following
complete state vector
\begin{equation} \label{eq:full_state}
    \boldsymbol{x} =
    \begin{bmatrix}
        \boldsymbol{x}_b     &
        \boldsymbol{x}_{I_1} & ... &
        \boldsymbol{x}_{I_N}
    \end{bmatrix}
\end{equation}

\subsection{Filter Prediction}
Many EKF formulations use \ac{IMU} measurements as the prediction step of the filter.
However, as this formulation uses multiple \acp{IMU} and makes no assumptions on the
synchronization or consistency of the measurements, the \ac{IMU} measurements are all
used as updates. Therefore, the following simpler propagation equations used are
\begin{equation}
    \begin{array}{lcl}
        \Dot{\boldsymbol{p}} = \boldsymbol{v} & ~ & \prescript{G}{B}{\Dot{q}} = \frac{1}{2} \Omega(\boldsymbol{\omega})\prescript{G}{B}{q} \\
        \Dot{\boldsymbol{v}} = \boldsymbol{a} & ~ & \Dot{\boldsymbol{\omega}} = \boldsymbol{\alpha}                                        \\
        \Dot{\boldsymbol{a}} = \boldsymbol{0} & ~ & \Dot{\boldsymbol{\alpha}} = \boldsymbol{0}
    \end{array}
\end{equation}
where
\begin{equation}
    \Omega(\boldsymbol{\omega}) =
    \begin{bmatrix}
        - \crossmat{\boldsymbol{\omega}} & \boldsymbol{\omega} \\
        \boldsymbol{\omega}^T            & 0
    \end{bmatrix}
\end{equation}
and the cross product matrix is defined as
\begin{equation}
    \crossmat{\boldsymbol{\omega}} =
    \begin{bmatrix}
        0         & -\omega_z & \omega_y  \\
        \omega_z  & 0         & -\omega_x \\
        -\omega_y & \omega_x  & 0
    \end{bmatrix}
\end{equation}
As such, the only nonlinear portion of the prediction step is the evolution
of the body's orientation. The state transition function is defined as
\begin{equation}
    \boldsymbol{f}
    \left(
    \HatBold{x}^-_{k-1}
    \right) =
    \left[
        \begin{array}{ccl}
            \HatBold{p}^-_{k-1}         & +       & \HatBold{v}^-_{k-1} \Delta t                                                              \\
            \HatBold{v}^-_{k-1}         & +       & (\HatBold{a}^-_{k-1} - \mathcal{C}(\prescript{G}{B}{q}^-_{k-1}) \boldsymbol{g} ) \Delta t \\
            \HatBold{a}^-_{k-1}         &         &                                                                                           \\
            \prescript{G}{B}{q}^-_{k-1} & \otimes & q(\HatBold{\omega}_k \Delta t)                                                            \\
            \HatBold{\omega}^-_{k-1}    & +       & \HatBold{\alpha}^-_{k-1} \Delta t                                                         \\
            \HatBold{\alpha}^-_{k-1}    &         &                                                                                           \\
            \HatBold{x}_{I_0}           &         &                                                                                           \\
            \vdots                      &         &                                                                                           \\
            \HatBold{x}_{I_N}
        \end{array}
        \right]
\end{equation}
where $\mathcal{C}(q)$ is the rotation matrix of quaternion $q$,
$\boldsymbol{g}$ is the gravity vector,
and $\otimes$ is quaternion multiplication.
By taking the partial derivatives, the linearized state transition matrix
$\boldsymbol{F}$, is found to be
\begin{equation}
    \boldsymbol{F}_k =
    \begin{bmatrix}
        \boldsymbol{F}_{pp} & \boldsymbol{F}_{po} & \zeros{9}{6N} \\
        \zeros{9}{9}        & \boldsymbol{F}_{oo} & \zeros{9}{6N} \\
        \zeros{6N}{9}       & \zeros{6N}{9}       & \eye{6N}      \\
    \end{bmatrix}
\end{equation}
Where
$\eye{n}$ is an identity matrix of size $n \times n$,
$\zeros{m}{n}$ is an zero matrix of size $m \times n$, and
the state transition matrices for position, orientation, and the correlation
terms, $\boldsymbol{F}_{pp}$, $\boldsymbol{F}_{oo}$, and $\boldsymbol{F}_{po}$
respectively are
\begin{align}
    \boldsymbol{F}_{pp} = &
    \begin{bmatrix}
        \boldsymbol{I}_{3} & \boldsymbol{I}_{3} \Delta t & \zeros{3}{3}                \\
        \zeros{3}{3}       & \boldsymbol{I}_{3}          & \boldsymbol{I}_{3} \Delta t \\
        \zeros{3}{3}       & \zeros{3}{3}                & \boldsymbol{I}_{3}
    \end{bmatrix}                          \\
    \boldsymbol{F}_{oo} = &
    \begin{bmatrix}
        \mathcal{C}(q(\HatBold{\omega}_k \Delta t)) & \boldsymbol{I}_{3} \Delta t & \zeros{3}{3}                \\
        \zeros{3}{3}                                & \boldsymbol{I}_{3}          & \boldsymbol{I}_{3} \Delta t \\
        \zeros{3}{3}                                & \zeros{3}{3}                & \boldsymbol{I}_{3}
    \end{bmatrix} \\
    \boldsymbol{F}_{po} = &
    \begin{bmatrix}
        \crossmat{\mathcal{C}(\prescript{G}{B}{q}^-_{k-1}) \boldsymbol{g} \Delta t} & \zeros{3}{6} \\
        \zeros{6}{3}                                                                & \zeros{6}{6}
    \end{bmatrix}
\end{align}
This state transition matrix is then used to predict the state and covariance
\begin{equation}
    \HatBold{x}^-_{k} =
    \boldsymbol{f}(\HatBold{x}^-_{k-1})
\end{equation}
\begin{equation}
    \boldsymbol{P}^-_{k} =
    \boldsymbol{F}_k
    \boldsymbol{P}^-_{k-1}
    \boldsymbol{F}^T_k +
    \boldsymbol{F}_k
    \boldsymbol{Q}_k
    \boldsymbol{F}_k^T
\end{equation}
where $\boldsymbol{Q}$ is defined as
\begin{equation}
    \boldsymbol{Q} =
    \left[
        \begin{array}{c:ccc}
            \boldsymbol{Q}_b &                      & \zeros{18}{12N} &                      \\ \hdashline
                             & \boldsymbol{Q}_{I_0} & \cdots          & \zeros{12}{12}       \\
            \zeros{12N}{18}  & \vdots               & \ddots          & \vdots               \\
                             & \zeros{12}{12}       & \cdots          & \boldsymbol{Q}_{I_N}
        \end{array}
        \right]
\end{equation}
where the body process noise $\boldsymbol{Q}_b$ is
\begin{equation}
    \boldsymbol{Q}_b =
    \begin{bmatrix}
        \zeros{6}{6} & \zeros{6}{3}     & \zeros{6}{3} & \zeros{6}{3}          & \zeros{6}{3} \\
        \zeros{3}{6} & \boldsymbol{Q}_a & \zeros{3}{3} & \zeros{3}{3}          & \zeros{3}{3} \\
        \zeros{3}{6} & \zeros{3}{3}     & \zeros{3}{3} & \zeros{3}{3}          & \zeros{3}{3} \\
        \zeros{3}{6} & \zeros{3}{3}     & \zeros{3}{3} & \boldsymbol{Q}_\omega & \zeros{3}{3} \\
        \zeros{3}{6} & \zeros{3}{3}     & \zeros{3}{3} & \zeros{3}{3}          & \zeros{3}{3}
    \end{bmatrix}
\end{equation}
The \ac{IMU} process noise matrices $\boldsymbol{Q}_{I}$ are zero in
the non-calibration case and the acceleration and angular rate noise matrices
$\boldsymbol{Q}_a$ and $\boldsymbol{Q}_\omega$ are computed offline.

\subsection{IMU Non-Calibration Update} \label{sec:udpate}

The \ac{IMU} measurement $\boldsymbol{z}$ is comprised of the measured acceleration
$\boldsymbol{a}_m$ and angular rate $\boldsymbol{\omega}_m$
\begin{equation}
    \boldsymbol{z} =
    \begin{bmatrix}
        \boldsymbol{a}_m \\
        \boldsymbol{\omega}_m
    \end{bmatrix}
\end{equation}
and the nonlinear measurement function $\boldsymbol{h}$ is
\begin{equation} \label{eq:pred_measurement}
    \begin{split}
        &\boldsymbol{h}(\HatBold{x}_{b}) = \\
        &\begin{bmatrix}
            \mathcal{C}(\prescript{B}{I_i}{q})^T
            \left(
            \mathcal{C}(\prescript{G}{B}{q})^T \HatBold{a} +
            \HatBold{\alpha} \times \prescript{B}{}{\HatBold{p}}_{I_i} +
            \HatBold{\omega} \times \HatBold{\omega} \times \prescript{B}{}{\HatBold{p}}_{I_i}
            \right)
            \\
            \mathcal{C}(\prescript{B}{I_i}{q})^T
            \mathcal{C}(\prescript{G}{B}{q})^T
            \HatBold{\omega}
        \end{bmatrix}
    \end{split}
\end{equation}
Therefore, the linearized observation matrix is
\begin{equation} \label{eq:obs_matrix}
    \boldsymbol{H} =
    \begin{bmatrix}
        \boldsymbol{H}_b     &
        \boldsymbol{H}_{i_0} &
        \cdots               &
        \boldsymbol{H}_{i_N}
    \end{bmatrix}
\end{equation}
where the \ac{IMU} calibration state Jacobians are zero for the non-calibration
case, and the body Jacobian $\boldsymbol{H}_b$ is
\begin{equation}
    \boldsymbol{H}_b =
    \begin{bmatrix}
        \zeros{3}{3}                                                              &
        \zeros{3}{3}                                                              &
        \frac{\partial \boldsymbol{a}_m}{\partial \hat{\boldsymbol{a}}}           &
        \zeros{3}{3}                                                              &
        \frac{\partial \boldsymbol{a}_m}{\partial \hat{\boldsymbol{\omega}}}      &
        \frac{\partial \boldsymbol{a}_m}{\partial \hat{\boldsymbol{\alpha}}}        \\
        \zeros{3}{3}                                                              &
        \zeros{3}{3}                                                              &
        \zeros{3}{3}                                                              &
        \zeros{3}{3}                                                              &
        \frac{\partial \boldsymbol{\omega}_m}{\partial \hat{\boldsymbol{\omega}}} &
        \zeros{3}{3}
    \end{bmatrix}
\end{equation}
Using the derivatives of a rotation inverse and coordinate map from
\cite{Bloesch2016APO}, the measurement Jacobians are as follows
\begin{align}
    \frac{\partial \boldsymbol{a}_m}{\partial \hat{\boldsymbol{a}}}           & =
    \mathcal{C}(\prescript{B}{I_i}{q})^T
    \mathcal{C}(\prescript{G}{B}{q})^T
    \\
    \frac{\partial \boldsymbol{a}_m}{\partial \hat{\boldsymbol{\omega}}}      & =
    \mathcal{C}(\prescript{B}{I_i}{q})^T
    \left(
    \crossmat{\hat{\boldsymbol{\omega}}}
    \crossmat{\prescript{B}{}{\HatBold{p}}_{I_i}}^T +
    \crossmat{\hat{\boldsymbol{\omega}} \times \prescript{B}{}{\HatBold{p}}_{I_i}}^T
    \right)
    \\
    \frac{\partial \boldsymbol{a}_m}{\partial \hat{\boldsymbol{\alpha}}}      & =
    \mathcal{C}(\prescript{B}{I_i}{q})^T
    \crossmat{\prescript{B}{}{\HatBold{p}}_{I_i}}^T
    \\
    \frac{\partial \boldsymbol{\omega}_m}{\partial \hat{\boldsymbol{\omega}}} & =
    \crossmat{
        \mathcal{C}(\prescript{B}{I_i}{q})^T
        \mathcal{C}(\prescript{G}{B}{q})^T
        \HatBold{\omega}
    }
\end{align}
With the linearized observation matrix, it is possible to perform a state
update using the \ac{EKF} formulation
\begin{equation} \label{eq:Kalman_gain}
    \boldsymbol{K}_k =
    \boldsymbol{P}_{k}^-
    \boldsymbol{H}_{k}^T
    \left(
    \boldsymbol{H}_{k}
    \boldsymbol{P}_{k}^-
    \boldsymbol{H}_{k}^T +
    \boldsymbol{R}_{k}
    \right)^{-1}
\end{equation}
\begin{equation} \label{eq:Kalman_state}
    \HatBold{x}^+_{k} =
    \HatBold{x}^-_{k} +
    \boldsymbol{K}_k
    \left(
    \boldsymbol{z}_k  - \boldsymbol{h}(\HatBold{x}^-_{k})
    \right)
\end{equation}
\begin{equation} \label{eq:Kalman_cov}
    \boldsymbol{P}^+_{k} =
    (\boldsymbol{I} - \boldsymbol{K}_k \boldsymbol{H}_k)
    \boldsymbol{P}^-_{k}
    (\boldsymbol{I} - \boldsymbol{K}_k \boldsymbol{H}_k)^T + \boldsymbol{K}_k \boldsymbol{R}_k \boldsymbol{K}_k ^T
\end{equation}

\subsection{IMU Calibration Update} \label{sec:cal_udpate}

For the case where the \ac{IMU} extrinsic calibrations are not extremely
well-known, it is desirable to produce an online estimate of these values.
This is especially true when considering the intrinsic biases of the \ac{IMU}
measurement, which drift with time and therefore cannot be precisely calibrated
significantly before measurements are taken. If there are exteroceptive
measurements, such as images in an \ac{MSCKF} framework, the calibration states
are observable for all \acp{IMU}. To perform this online calibration,
the state is extended as in \cref{eq:IMU_state,eq:full_state}. The process noise
matrices for the calibration states of \ac{IMU} $i$ is
\begin{equation}
    \boldsymbol{Q}_{I_i}=
    \begin{bmatrix}
        \zeros{6}{6} & \zeros{6}{3}         & \zeros{6}{3}              \\
        \zeros{3}{6} & \boldsymbol{Q}_{b_a} & \zeros{3}{3}              \\
        \zeros{3}{6} & \zeros{3}{3}         & \boldsymbol{Q}_{b_\omega}
    \end{bmatrix}
\end{equation}
where the accelerometer and gyroscope bias noise matrices,
$\boldsymbol{Q}_a$ and $\boldsymbol{Q}_\omega$, are computed offline. These
definitions are used in the prediction step of the filter. For the update step,
the linearized \ac{IMU} observation matrix for the \textit{ith} \ac{IMU}'s calibration states is
\begin{equation}
    \boldsymbol{H}_i =
    \begin{bmatrix}
        \frac{\partial \boldsymbol{a}_m}{\partial \prescript{B}{}{\HatBold{p}}_{I_i}} &
        \frac{\partial \boldsymbol{a}_m}{\partial \prescript{B}{I_i}{\Hat{q}}}        &
        \eye{3}                                                                       &
        \zeros{3}{3}                                                                    \\
        \zeros{3}{3}                                                                  &
        \frac{\partial\boldsymbol{\omega}_m }{\partial \prescript{B}{I_i}{\Hat{q}}}   &
        \zeros{3}{3}                                                                  &
        \eye{3}
    \end{bmatrix}
\end{equation}
where the measurement Jacobians are
\begin{align}
    \frac{\partial \boldsymbol{a}_m}{\partial \prescript{B}{}{\HatBold{p}}_{I_i}} & =
    \quatRot{\prescript{B}{I_i}{\Hat{q}}}^T
    \left(
    \crossmat{\boldsymbol{\alpha}} + \crossmat{\HatBold{\omega}} \crossmat{\HatBold{\omega}}
    \right)                                                                           \\
    \frac{\partial \boldsymbol{a}_m}{\partial \prescript{B}{I_i}{\Hat{q}}}        & =
    \crossmat{
        \quatRot{\prescript{B}{I_i}{\Hat{q}}}^T
        \left(
        (\HatBold{\alpha} + \HatBold{\omega} \times \HatBold{\omega} ) \times \prescript{B}{}{p}_{I_i} +
        \quatRot{\prescript{G}{B}{\Hat{q}}}^T
        \HatBold{a}
        \right)
    }                                                                                 \\
    \frac{\partial\boldsymbol{\omega}_m }{\partial \prescript{B}{I_i}{\Hat{q}}}   & =
    \crossmat{
        \quatRot{\prescript{B}{I_i}{\Hat{q}}}^T
        \quatRot{\prescript{G}{B}{\Hat{q}}}^T
        \HatBold{\omega}
    }
\end{align}
This calibration state observation matrix is calculated for each \ac{IMU} and
combined with the body state observation matrix as in \cref{eq:obs_matrix}. The
rest of the Kalman update shown in
\cref{eq:Kalman_gain,eq:Kalman_state,eq:Kalman_cov} remain the same.

\subsection{Calibration State Observability} \label{sec:obs}

Unfortunately, the filter states as described thus far are not observable with
the \ac{IMU} measurements alone. This can be confirmed using the typical methods
\begin{equation}
    \text{rank} (O) = (18+12N)-6
\end{equation}
where the observability matrix $O$ is
\begin{equation}
    O =
    \begin{bmatrix}
        \boldsymbol{H}                        \\
        \boldsymbol{H}\boldsymbol{F}          \\
        \boldsymbol{H}\boldsymbol{F}^{2}      \\
        \vdots                                \\
        \boldsymbol{H}\boldsymbol{F}^{18+12N} \\
    \end{bmatrix}
\end{equation}
Because the observability matrix rank is six degrees of freedom fewer than the
state size, there are six states that are not observable using \ac{IMU}
measurements alone. In practice, this means that intrinsic biases cannot be
estimated for all \acp{IMU}, but rather at least one \ac{IMU} would need to have
a constant and pre-calibrated bias estimate. As this is not reliably possible,
it is necessary to pair this multi-IMU filter with additional exteroceptive
measurements, such as feature tracks within an \ac{MSCKF} framework, to make all \ac{IMU}
calibration states observable. Therefore, utilizing the proposed filter
methodology alongside exteroceptive measurements produces online estimates
of the \ac{IMU} intrinsic biases and extrinsic parameters in addition to
proprioceptive estimates of the body motion. Note, in the remainder of this
work, a camera and \ac{MSCKF} framework is utilized.

\section{Simulation Results} \label{sec:sim}

For initial validation of the calibration filter's stability and robustness, a
Monte Carlo simulation was developed that randomly generates 6 degree of freedom
trajectories and randomly assigns errors to the intrinsic and extrinsic calibration estimates.
This open source simulation and evaluation software is available at \cite{ekf-cal}.
\begin{figure}
    \centering
    \includegraphics[width = 0.9\linewidth]{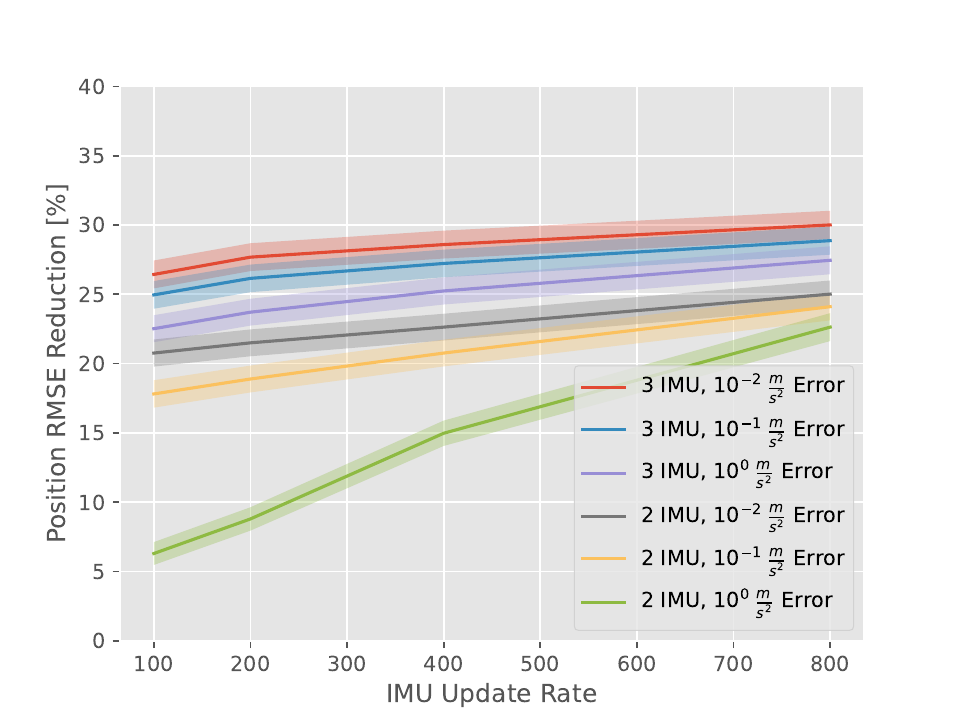}
    \caption{Reduction of VIO position RMS errors with various \ac{IMU} counts
        and acceleration error values without online calibration compared to a
        single VN-300. A 95\% confidence interval is shown after taking 1000 samples with
        100 second run times.}
    \label{fig:multi_imu_no_cal}
\end{figure}
\begin{figure}
    \centering
    \includegraphics[width = 0.9\linewidth]{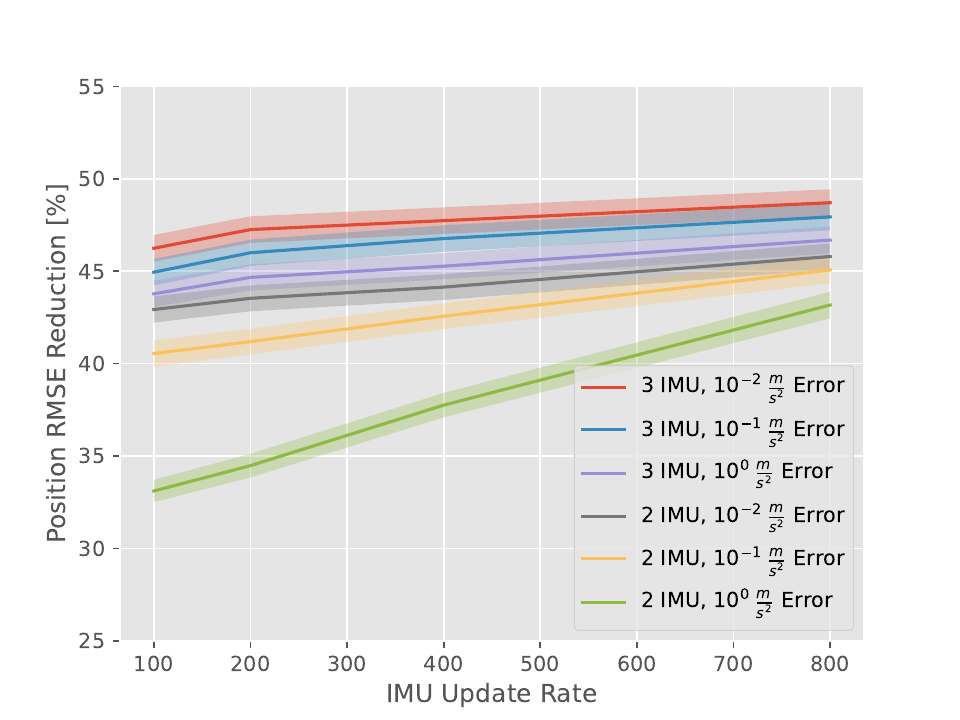}
    \caption{Reduction of VIO position RMS errors with various \ac{IMU} counts
        and acceleration error values with online calibration compared to a single VN-300 when
        there are initialization errors with standard deviations of $5^\circ$
        and 20 mm for orientation and position, respectively.
        A 95\% confidence interval is shown after taking 1000 samples with
        100 second run times.}
    \label{fig:multi_imu_cal}
\end{figure}
\cref{fig:multi_imu_no_cal,fig:multi_imu_cal} show the relative improvement from
utilizing multiple \acp{IMU} in the proposed update filter scheme over simply
using a single \ac{IMU} as the filter predictor. \cref{fig:multi_imu_no_cal}
outlines the simulated improvements when utilizing the non-calibration update
schema outlined in \cref{sec:udpate}, whereas \cref{fig:multi_imu_cal}
outlines the simulated improvements when utilizing the calibration update
schema outlined in \cref{sec:cal_udpate} with the addition of initialization
error with standard deviations of $5^\circ$ and 20 mm for \ac{IMU} orientation
and position within the body frame. Both sets of scenarios show a general trend
of increasing VIO accuracy as the number of \acp{IMU} increases and the
measurement error of each \ac{IMU} decreases. However, the calibration filter
comparisons with induced errors show much larger reductions in relative error.
This is most likely due to the impact of drift on a VIO without loop closure,
such as the \ac{MSCKF} used for visual processing. As such, the values of
relative improvement must only be used in context, as the absolute values can
vary drastically due to differences in run time. The key result is the
continued trend of decreasing RMS position error as the quality of \ac{IMU}
improves or the quantity of \ac{IMU} increases. Interestingly, these simulation
suggest the introduction of two low-rate, low-quality sensors improves the
system's performance as much as a single high-rate, high-quality sensor. Given
a system that is processor-limited therefore, it may be most beneficial to add
two independent, low rate \acp{IMU} to an existing system than a single
higher rate sensor.

\begin{figure}
    \centering
    \includegraphics[width = 0.9\linewidth]{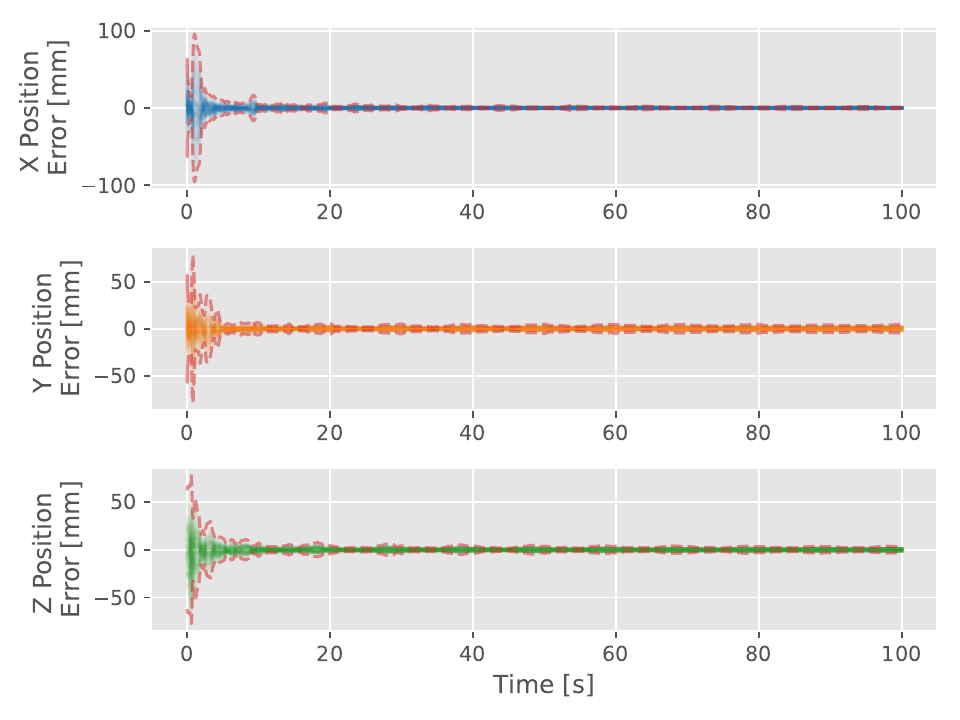}
    \caption{\ac{IMU} extrinsic position parameter filter convergence given a 20
        mm standard deviation initial error using a VN-300 and VN-100. 1000 samples were taken with
        100 second run times with the $3\sigma$ standard error shown in red.}
    \label{fig:imu_pos_cal}
\end{figure}

\begin{figure}
    \centering
    \includegraphics[width = 0.9\linewidth]{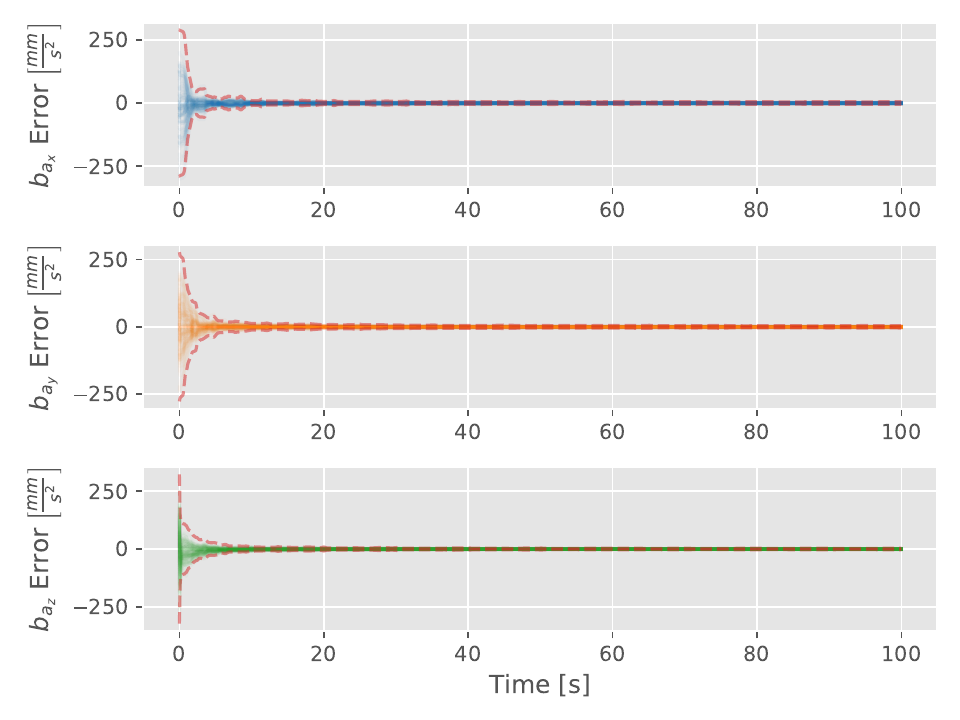}
    \caption{\ac{IMU} intrinsic accelerometer bias convergence
        given a 0.1 $\frac{m}{s^2}$ standard deviation initial error using a
        VN-300 and VN-100. 1000 samples were taken with
        100 second run times with the $3\sigma$ standard error shown in red.}
    \label{fig:imu_acc_bias}
\end{figure}

\begin{figure}
    \centering
    \includegraphics[width = 0.9\linewidth]{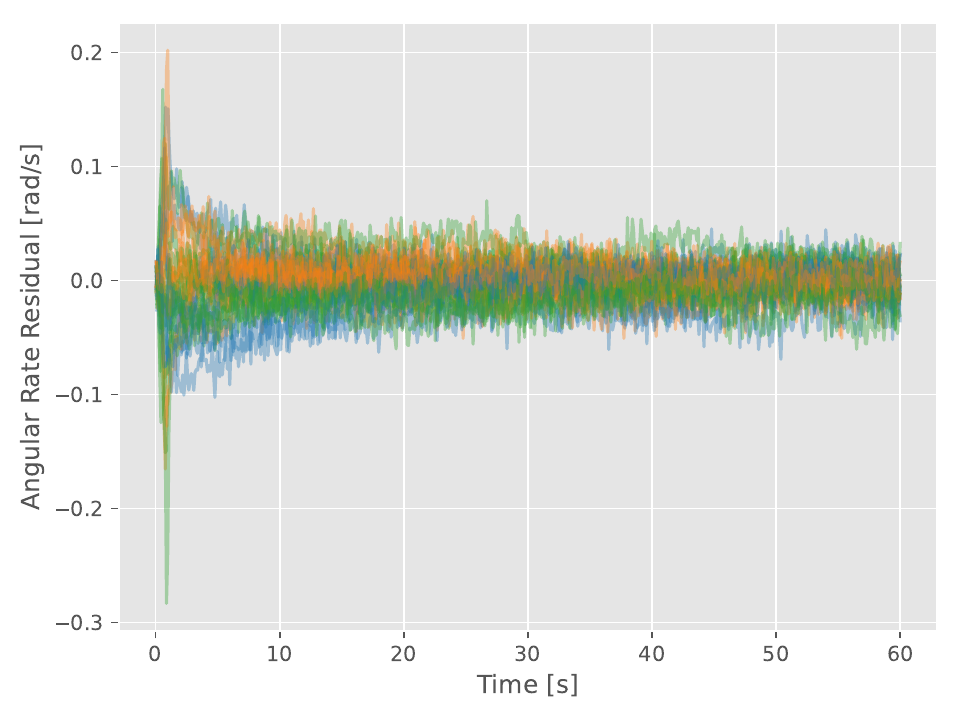}
    \caption{Angular rate measurement residual convergence given a $5^\circ$
        standard deviation initial angular error using a VN-300 and VN-100.
        100 samples were taken with 60 second run times.
    }
    \label{fig:imu_omg_resid}
\end{figure}

Additionally, the convergence of the filter on each calibration parameter was
verified using Monte Carlo simulations. An experiment was run where initial
errors in position and orientation were applied with 20 mm and $5^\circ$
standard deviations, respectively. The convergence of each calibration parameter
was confirmed with final mean and standard deviations of each error outlined in
\cref{tab:sim_results} and example convergence plots for the second \ac{IMU}
extrinsic position, intrinsic accelerometer bias, and gyroscope residuals are shown in
\cref{fig:imu_pos_cal}, \cref{fig:imu_acc_bias}, and \cref{fig:imu_omg_resid}
respectively.

\begin{table}
    \centering
    \caption{
        Summary of \ac{IMU} intrinsic errors across 1000 samples using a
        VN-300 and VN-100 \ac{IMU}
    }
    \label{tab:sim_results}
    \begin{tabular}{l c c c}
        Parameter          & Unit             & Mean & Standard Deviation \\
        \hline
        Position           & mm               & 0.10 & 1.1                \\
        Orientation        & mrad             & 0.20 & 0.80               \\
        Accelerometer Bias & $\frac{mm}{s^2}$ & 0.30 & 1.5                \\
        Gyroscope Bias     & $\frac{mrad}{s}$ & 0.18 & 0.96
    \end{tabular}
\end{table}

\begin{table}
    \centering
    \caption{Summary of the noise densities for the \acp{IMU} used in simulation and experiments}
    \label{tab:noise_densities}
    \begin{tabular}{l c c}
        \ac{IMU} & Accelerometer $\left[\frac{mg}{\sqrt{Hz}}\right]$ & Gyroscope $\left[\frac{mrad}{s \sqrt{Hz}}\right]$ \\
        \hline
        VN-300   & 0.14                                              & 0.061                                             \\
        VN-100   & 0.14                                              & 0.061                                             \\
        DETA10   & 40                                                & 0.049
    \end{tabular}
\end{table}

\section{Experimental Results}

\begin{figure}
    \centering
    \includegraphics[width = 0.9\linewidth]{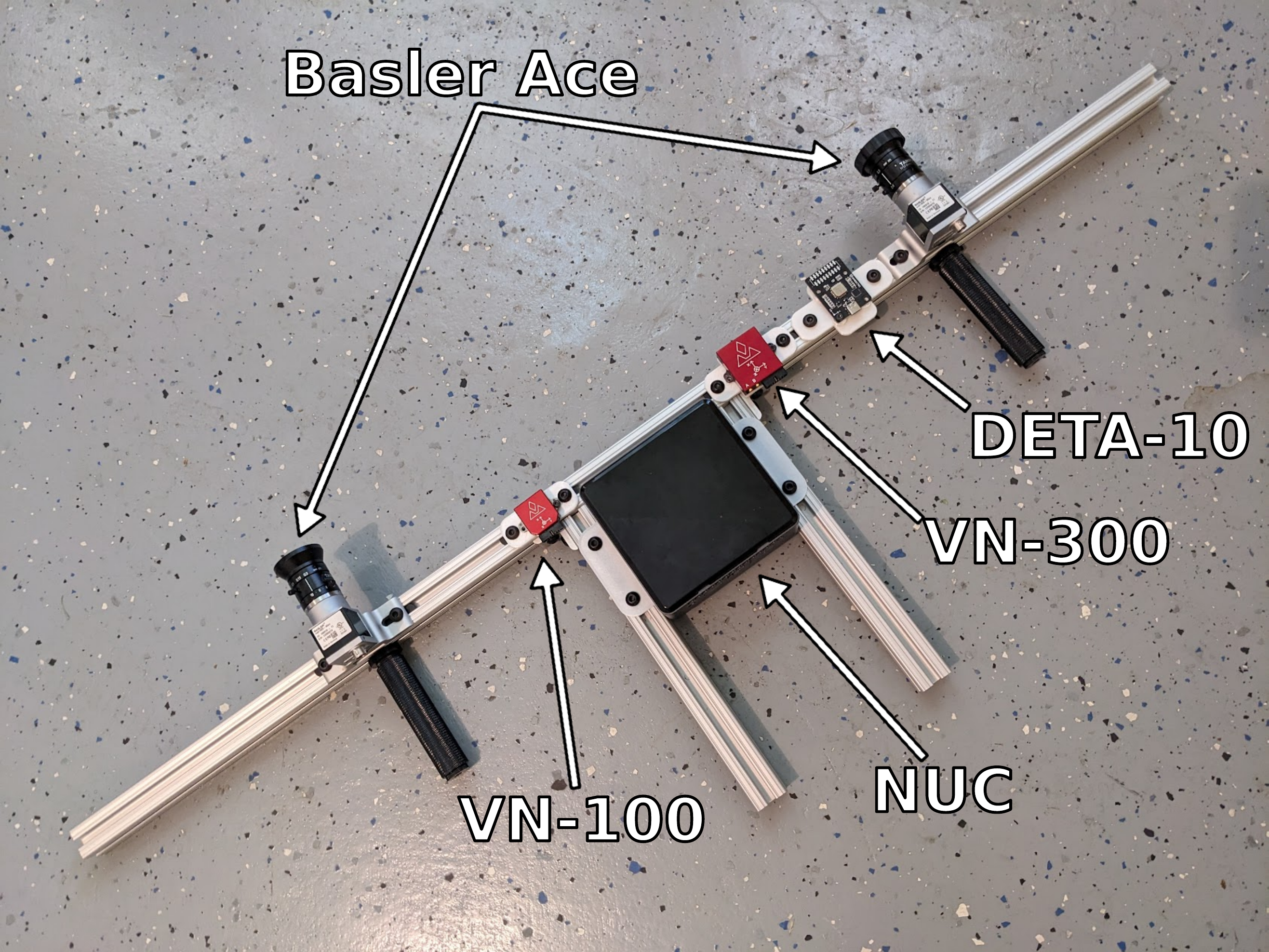}
    \caption{Experimental rig showing two Basler Ace acA1920-40uc USB 3.0
        cameras, a Vectornav VN-300, Vectornav VN-100, and a FDI Systems DETA-10
        run using a NUC8i7BEH.}
    \label{fig:rig}
\end{figure}

To further validate the efficacy of the proposed multi-IMU filtering method, experiments using three \acp{IMU}
and a single camera were
conducted while running the multi-IMU \ac{MSCKF} filter through an
open-sourced ROS node which is available at \cite{ekf-cal} along with the
simulation used in \cref{sec:sim}.
The base \ac{IMU} in the experiment was a Vectornav VN-300 running at 400 Hz.
This sensor was chosen in particular as it uses \ac{GPS} measurements for an \ac{INS}
solution which was used as an estimate of ground
truth for body angular rates and accelerations while the raw \ac{IMU}
measurements are still available separately. The two additional \acp{IMU} were a
Vectornav VN-100 running at 100 to 800 Hz and a FDI Systems DETA-10 optionally
running at 100 Hz.

\begin{figure}
    \centering
    \includegraphics[width = 0.9\linewidth]{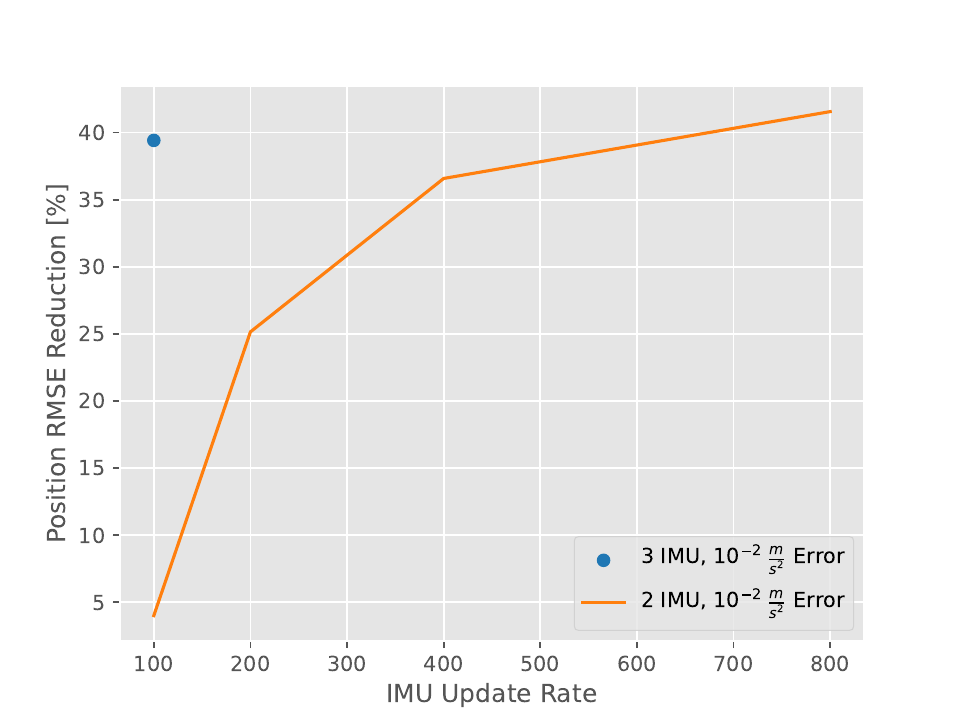}
    \caption{Reduction of VIO position RMS drift over 60 second run time with
        various \ac{IMU} counts and acceleration error values with online
        calibration compared to a single VN-300.}
    \label{fig:exp_results}
\end{figure}

Using the \ac{GPS}-aided \ac{INS} output of the VN-300 as an estimate of ground truth,
it is possible to calculate the \ac{RPE} using the method outlined in \cite{Prokhorov}.
The reduction of \ac{RPE} was calculated while running the same trajectory with various sensor
update rates and combinations as compared to running a single IMU as the
prediction step of an \ac{MSCKF}. Given the errors of each \ac{IMU} are fixed,
and the maximum achieved rate of the DETA-10 was 100 Hz, the resulting summary
in \cref{fig:exp_results} is limited in its scope. However, the trend that is
shown with increasing the rate of the secondary IMU promisingly appears to match
what was observed in simulation. Moreover, as in simulation,
it appears that adding more stochastically independent sensors may have a
greater impact on the reduction of VIO drift than adding a single high-rate
sensor, depending on the update rates and standard errors involved.

\begin{figure}
    \centering
    \includegraphics[width = 0.9\linewidth]{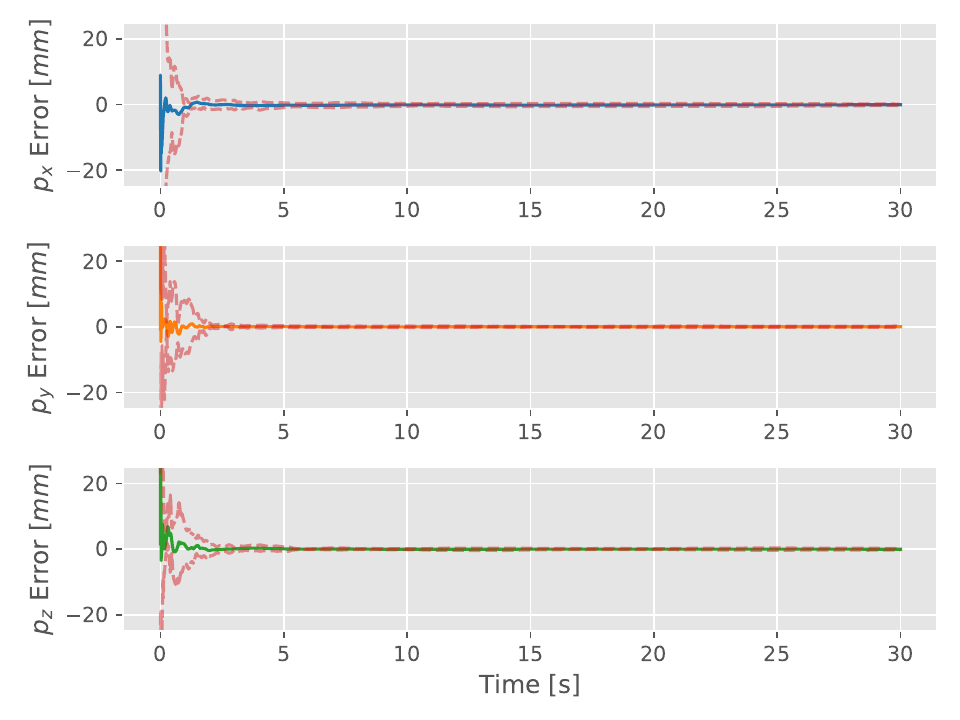}
    \caption{Accelerometer extrinsic position error as compared to Kalibr over 60 seconds
        using a 40 hz 640 x 400 camera, a VN-300, and a VN-100.}
    \label{fig:exp_pos_results}
\end{figure}

\begin{figure}
    \centering
    \includegraphics[width = 0.9\linewidth]{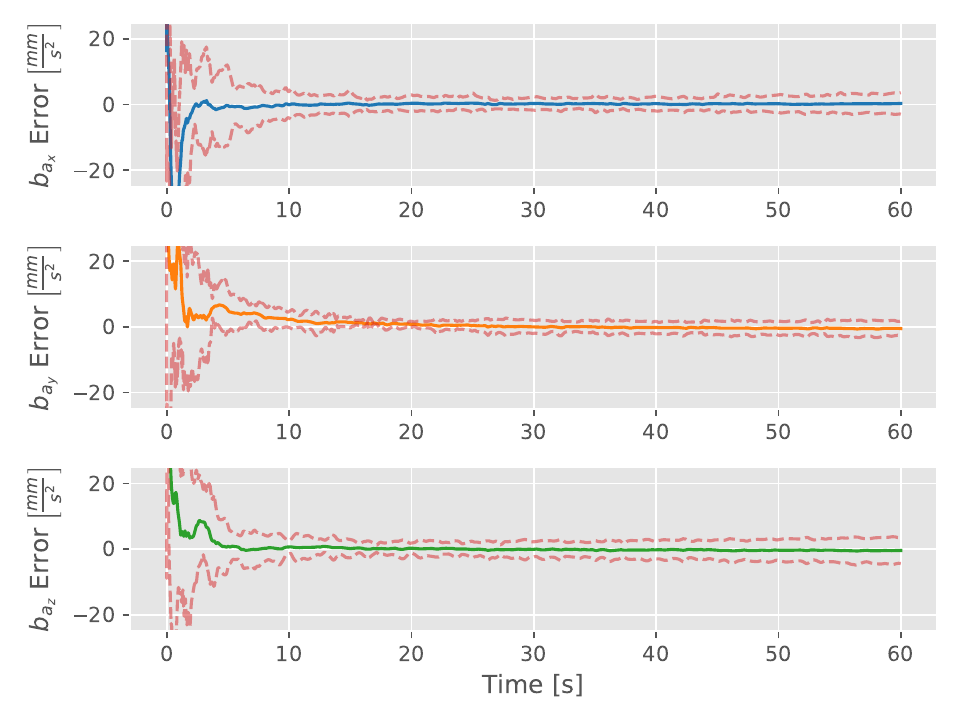}
    \caption{Accelerometer intrinsic bias error as compared to Kalibr over 60 seconds
        using a 40 hz 640 x 400 camera, a VN-300, and a VN-100.}
    \label{fig:exp_bias_results}
\end{figure}

Additionally, the experimental results of the filter were compared to the output
produced by Kalibr, a state-of-the-art nonlinear optimizer \cite{Rehder}. The
final errors, compared to Kalibr for a 60 second sensor sequence with a
calibration checkerboard visible produced very similar extrinsic parameters as
outlined in \cref{tab:kalibr_comparison} with an example of the accelerometer extrinsic position
and intrinsic bias parameters converging on the Kalibr value shown in
\cref{fig:exp_pos_results,fig:exp_bias_results}.
These low errors in nominal lab environments further validate the filter
stability and overall accuracy.

\begin{table}
    \centering
    \caption{Extrinsic errors as compared to Kalibr over 60 seconds
        using a 40 hz 640 x 400 camera, a VN-300, and a VN-100.}
    \label{tab:kalibr_comparison}
    \begin{tabular}{l c c c}
        Extrinsic Parameter & Unit             & Error \\
        \hline
        Position Offset     & mm               & 0.96  \\
        Orientation Offset  & mrad             & 0.52  \\
        Accelerometer Bias  & $\frac{mm}{s^2}$ & 0.86  \\
        Gyroscope Bias      & $\frac{mrad}{s}$ & 1.5
    \end{tabular}
\end{table}

\section{Conclusions}
This work has outlined a novel centralized \ac{EKF}-based method for the
combination and calibration of multiple unsynchronized \acp{IMU}. Unlike
existing frameworks, the proposed filter does not rely on the imposition of
geometric constraints and is truly online in its estimate of intrinsic and
extrinsic calibration parameters. Additionally, the framework is flexible in the
update rate, quality, and error model used for each \ac{IMU} and has been
incorporated into an \ac{MSCKF} for \ac{VIO}. This combined system
has been validated for filter stability and accuracy both using a Monte Carlo simulation as
well as experimentally. In both simulations and experiments, using multiple
IMU measurement streams within the proposed filter framework
outperformed the use of a single IMU in a filter prediction step while also
producing consistent and accurate estimates of initial calibration errors.
In certain cases, adding multiple low-quality sensors outperformed adding a
single high-quality sensor, while being less computationally expensive.
Compared to current state-of-the-art optimizers, the filter produces similar
intrinsic and extrinsic calibration parameters for each of additional sensor.
Lastly, an open source repository has been proved containing both the online
estimator and the simulation used for Monte Carlo testing and evaluation.

\bibliographystyle{IEEEtran}
\bibliography{references.bib}

\begin{thebibliography}{10}
\providecommand{\url}[1]{#1}
\csname url@samestyle\endcsname
\providecommand{\newblock}{\relax}
\providecommand{\bibinfo}[2]{#2}
\providecommand{\BIBentrySTDinterwordspacing}{\spaceskip=0pt\relax}
\providecommand{\BIBentryALTinterwordstretchfactor}{4}
\providecommand{\BIBentryALTinterwordspacing}{\spaceskip=\fontdimen2\font plus
\BIBentryALTinterwordstretchfactor\fontdimen3\font minus
  \fontdimen4\font\relax}
\providecommand{\BIBforeignlanguage}[2]{{%
\expandafter\ifx\csname l@#1\endcsname\relax
\typeout{** WARNING: IEEEtran.bst: No hyphenation pattern has been}%
\typeout{** loaded for the language `#1'. Using the pattern for}%
\typeout{** the default language instead.}%
\else
\language=\csname l@#1\endcsname
\fi
#2}}
\providecommand{\BIBdecl}{\relax}
\BIBdecl

\bibitem{Faizullin}
M.~Faizullin and G.~Ferrer, ``Best axes composition: Multiple gyroscopes imu
  sensor fusion to reduce systematic error,'' in \emph{2021 European Conference
  on Mobile Robots (ECMR)}, 2021, pp. 1--7.

\bibitem{Patel}
U.~N. Patel and I.~A. Faruque, ``Multi-imu based alternate navigation
  frameworks: Performance and comparison for uas,'' \emph{IEEE Access},
  vol.~10, pp. 17\,565--17\,577, 2022.

\bibitem{Bancroft_2011}
\BIBentryALTinterwordspacing
J.~B. Bancroft and G.~Lachapelle, ``Data fusion algorithms for multiple
  inertial measurement units,'' \emph{Sensors}, vol.~11, no.~7, pp. 6771--6798,
  2011. [Online]. Available: \url{https://www.mdpi.com/1424-8220/11/7/6771}
\BIBentrySTDinterwordspacing

\bibitem{Zhang}
M.~Zhang, X.~Xu, Y.~Chen, and M.~Li, ``A lightweight and accurate localization
  algorithm using multiple inertial measurement units,'' \emph{IEEE Robotics
  and Automation Letters}, vol.~5, no.~2, pp. 1508--1515, 2020.

\bibitem{Egidio}
E.~D'Amato, M.~Mattei, A.~Mele, I.~Notaro, and V.~Scordamaglia, ``Fault
  tolerant low cost imus for uavs,'' in \emph{2017 IEEE International Workshop
  on Measurement and Networking (M\&N)}, 2017, pp. 1--6.

\bibitem{Eckenhoff}
K.~Eckenhoff, P.~Geneva, and G.~Huang, ``Sensor-failure-resilient multi-imu
  visual-inertial navigation,'' in \emph{2019 International Conference on
  Robotics and Automation (ICRA)}, 2019, pp. 3542--3548.

\bibitem{Huang_2023}
H.~Huang, H.~Zhang, and L.~Jiang, ``An optimal fusion method of multiple
  inertial measurement units based on measurement noise variance estimation,''
  \emph{IEEE Sensors Journal}, vol.~23, no.~3, pp. 2693--2706, 2023.

\bibitem{Larey}
A.~Larey, E.~Aknin, and I.~Klein, ``Multiple inertial measurement units-an
  empirical study,'' \emph{IEEE Access}, vol.~8, pp. 75\,656--75\,665, 2020.

\bibitem{Carlson_federated}
N.~Carlson, ``Federated filter for fault-tolerant integrated navigation
  systems,'' in \emph{IEEE PLANS '88.,Position Location and Navigation
  Symposium, Record. 'Navigation into the 21st Century'.}, 1988, pp. 110--119.

\bibitem{Luo_2021}
Q.~Luo, X.~Yan, Z.~Zhou, C.~Wang, and C.~Hu, ``An integrated navigation and
  localization system,'' in \emph{2021 IEEE International Conference on Smart
  Internet of Things (SmartIoT)}, 2021, pp. 28--32.

\bibitem{Wang_2021}
L.~Wang, H.~Tang, T.~Zhang, Q.~Chen, J.~Shi, and X.~Niu, ``Improving the
  navigation performance of the mems imu array by precise calibration,''
  \emph{IEEE Sensors Journal}, vol.~21, no.~22, pp. 26\,050--26\,058, 2021.

\bibitem{Rehder}
J.~Rehder, J.~Nikolic, T.~Schneider, T.~Hinzmann, and R.~Siegwart, ``Extending
  kalibr: Calibrating the extrinsics of multiple imus and of individual axes,''
  in \emph{2016 IEEE International Conference on Robotics and Automation
  (ICRA)}, 2016, pp. 4304--4311.

\bibitem{Bancroft_2009}
J.~B. Bancroft, ``Multiple imu integration for vehicular navigation,'' in
  \emph{Proceedings of the 22nd International Technical Meeting of The
  Satellite Division of the Institute of Navigation (ION GNSS 2009)}, 2009, pp.
  1828--1840.

\bibitem{MIMCVINS}
K.~Eckenhoff, P.~Geneva, and G.~Huang, ``{MIMC-VINS}: A versatile and resilient
  multi-imu multi-camera visual-inertial navigation system,'' \emph{IEEE
  Transactions on Robotics}, vol.~37, no.~5, pp. 1360--1380, 2021.

\bibitem{MSCKF}
A.~I. Mourikis and S.~I. Roumeliotis, ``A multi-state constraint kalman filter
  for vision-aided inertial navigation,'' in \emph{Proceedings 2007 IEEE
  International Conference on Robotics and Automation}, 2007, pp. 3565--3572.

\bibitem{Bloesch2016APO}
\BIBentryALTinterwordspacing
M.~Bloesch, H.~Sommer, T.~Laidlow, M.~Burri, G.~N{\"u}tzi, P.~Fankhauser,
  D.~Bellicoso, C.~Gehring, S.~Leutenegger, M.~Hutter, and R.~Y. Siegwart, ``A
  primer on the differential calculus of 3d orientations,'' \emph{ArXiv}, vol.
  abs/1606.05285, 2016. [Online]. Available:
  \url{https://api.semanticscholar.org/CorpusID:2141}
\BIBentrySTDinterwordspacing

\bibitem{ekf-cal}
\BIBentryALTinterwordspacing
J.~Hartzer, ``ekf-cal: Extended kalman filter calibration and localization,''
  2023. [Online]. Available: \url{https://github.com/unmannedlab/ekf-cal}
\BIBentrySTDinterwordspacing

\bibitem{Prokhorov}
D.~Prokhorov, D.~Zhukov, O.~Barinova, K.~Anton, and A.~Vorontsova, ``Measuring
  robustness of visual slam,'' in \emph{2019 16th International Conference on
  Machine Vision Applications (MVA)}, 2019, pp. 1--6.

\end{thebibliography}

\end{document}